\documentclass{article}

\usepackage[preprint]{neurips_2023}

\usepackage[utf8]{inputenc} 
\usepackage[T1]{fontenc}    
\usepackage{hyperref}       
\usepackage{url}            
\usepackage{booktabs}       
\usepackage{amsfonts}       
\usepackage{nicefrac}       
\usepackage{microtype}      
\usepackage{xcolor}         
\usepackage{graphicx}
\usepackage{tabularx}

\usepackage{blindtext}
\newcommand\blfootnote[1]{%
\begingroup
\renewcommand\thefootnote{}\footnote{#1}%
\addtocounter{footnote}{-1}%
\endgroup
}

\title{EHRTutor: Enhancing Patient Understanding of Discharge Instructions}

\author{%
  Zihao Zhang\thanks{indicates equal contribution} $^{1}$, 
  Zonghai Yao\footnotemark[1] $^{1}$, 
  Huixue Zhou$^{2}$, 
  Feiyun ouyang$^{3}$, 
  Hong Yu$^{1,3}$
  \\
  University of Massachusetts, Amherst$^1$, University of Minnesota, Twin Cities$^2$\\
  University of Massachusetts, Lowell$^3$
  \\
  \texttt\ zihaozhang@umass.edu, zonghaiyao@umass.edu \\
  zhou1742@umn.edu, feiyun\_ouyang@uml.edu, hong\_yu@uml.edu 
  \\
}

\begin{document}

\maketitle

\begin{abstract}

            Large language models have shown success as a tutor in education in various fields. Educating patients about their clinical visits plays a pivotal role in patients' adherence to their treatment plans post-discharge. This paper presents EHRTutor, an innovative multi-component framework leveraging the Large Language Model (LLM) for patient education through conversational question-answering. EHRTutor first formulates questions pertaining to the electronic health record discharge instructions. It then educates the patient through conversation by administering each question as a test. Finally, it generates a summary at the end of the conversation. Evaluation results using LLMs and domain experts have shown a clear preference for EHRTutor over the baseline. Moreover, EHRTutor also offers a framework for generating synthetic patient education dialogues that can be used for future in-house system training.

\end{abstract}
\blfootnote{To appear in NeurIPS'23 Workshop on Generative AI for Education (GAIED)}

\section{Introduction}
 Studies have shown that patients have trouble understanding electronic health records (EHRs). Discharge instructions\citep{tutty2019complex} are a specific type of EHR notes where upon discharge, a provider presents and explains written instructions to the patient. Discharge instructions are important to patients as they provide critical information for patients to manage their care. However, discharge instructions are laden with medical terminology, which are difficult for patients to comprehend. Large Language Models (LLMs), such as GPT-3 and its successors, have emerged as powerful tools in various domains due to their
natural language processing capabilities\citep{brown2020language, openai2023gpt4}. LLMs have shown potential in healthcare communication \citep{yunxiang2023chatdoctor, zhang2023huatuogpt, wang2023umass_bionlp, Wang2023NoteChatAD} and decision-making \citep{BENABACHA2015570, Yang2023.10.26.23297629}. 
Therefore, we argue that LLMs can help patients understand their discharge instructions, and this can help improve self-managed care post-discharge.

\begin{figure*}[t] 
    \centering
    \includegraphics[width=10cm]{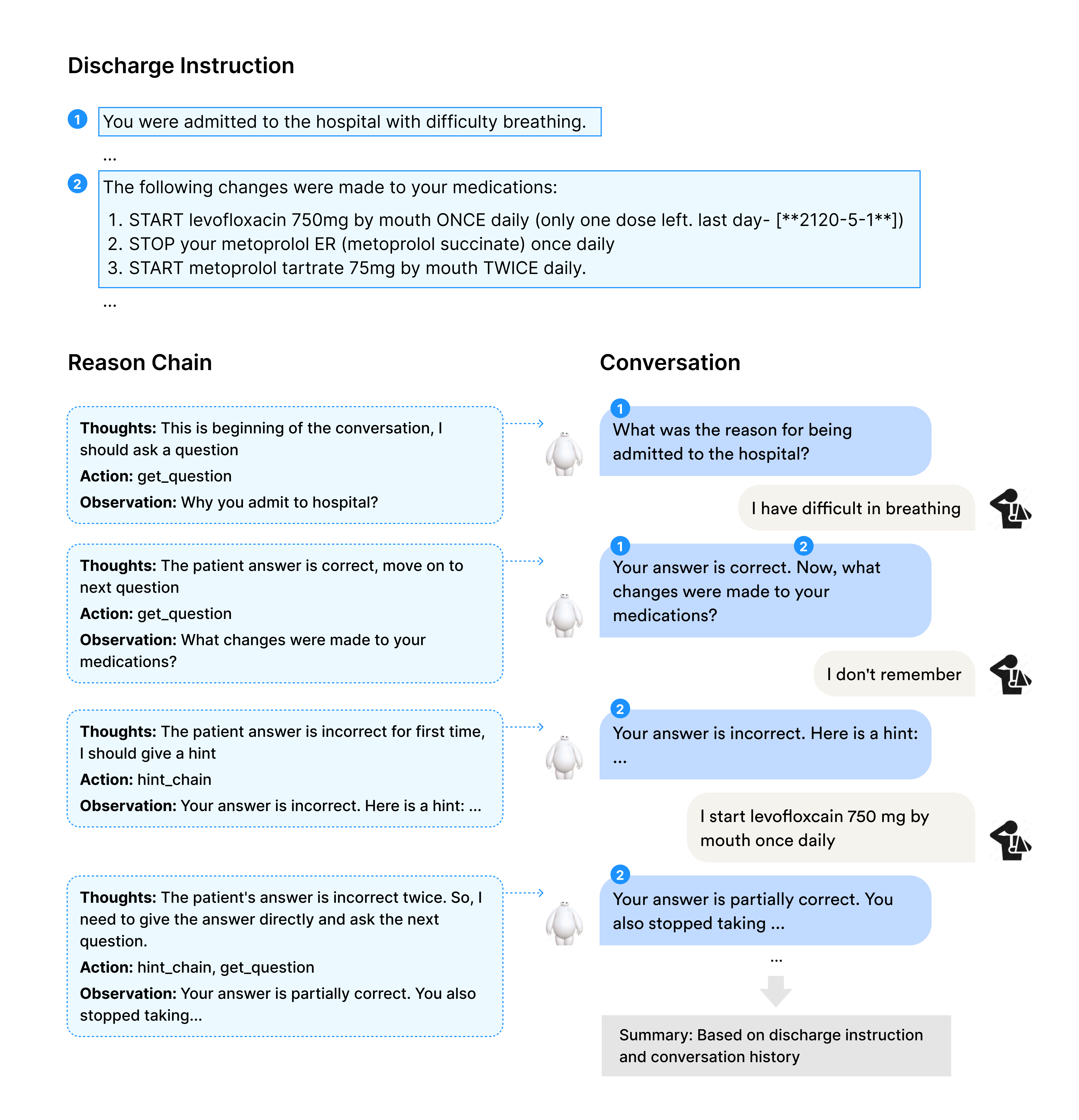}
    \caption{\textbf{Example for EHRTutor-patient interaction.} In this figure, we display a short example of a conversation between the EHRTutor and the patient. The reason chain is the decision-making process for the EHRTutor, which is not shown to the patient. The highlighted content in the discharge instruction is the answer to the corresponding question in the conversation. For example, the agent asked about the change of medications shown in the discharge instructions, but the patient did not remember. Then, the EHRTutor will give a hint based on the highlighted content.}
    \label{fig:conversation_example}
    \vspace{-3mm}
\end{figure*}

A recent study \citep{cai2023paniniqa} has shown a question-answering system called PaniniQA can help patients understand their discharge instructions. In that work, PaniniQA first generates questions using ChatGPT (GPT3.5) based on some model-predicted key points and then provides patients with the correct answers if patients' answers are incorrect. 
However, the quality of the questions generated by ChatGPT highly relied on their name entity recognition and relation extraction modules' output. These two modules are based on small language models and trained on a minimal number of data, resulting in lousy generalizability to real-world cases.
For example, we found some questions may be difficult for patients to understand. Some questions may not be relevant and, therefore, may not even have answers. 
Moreover, PaniniQA lacks a more natural way of communicating with patients: conversation. Instead, it provided the correct answers when patients' answers were incorrect. 
In contrast, EHRTutor generates questions using GPT4 with instruction prompting. The instructions were question templates formulated by domain experts. Examples of instructions are included in Table 1.
Secondly, as shown in Figure \ref{fig:conversation_example}, EHRTutor does not provide the correct answers directly. Instead, EHRTutor interacts with its patients by encouraging them to come up with the correct answers by providing them with hints. 
Studies have shown that this interactive question-answering helps tutors to be more engaged and to learn more in-depth \citep{andrini2016effectiveness, kuhlthau2015guided}. EHRTutor outputs a summary as the education feedback at the end of the conversation. The summary outlines the overall clinical content of a discharge instruction and highlights information important to the patients. The summary will help patients understand their discharge instructions \citep{metcalfe2007principles, patel2023chatgpt}.


To accomplish this, we split the EHRTutor pipeline into three modules: question generation, conversation, and summarization. The question generation module includes a template-based question generation chain followed by the question verification chain based on LLMs. The question generation module ensures the generated questions cover important information like medication usage, treatment data, etc. The conversation module is used to interact with patients. We implement the EHRTutor based on the Reasoning + Acting (ReAct) framework \citep{yao2023react}. The advantage of the ReAct is that it can prompt LLMs to produce verbal reasoning steps and actions for a given task. This helps the EHRTutor make more reasonable decisions on patient education through interaction. Once the conversation ends, the EHRTutor will send the discharge instructions and conversation history to the summarization module. The summarization module will provide the summary to the patient, including the key points of the discharge instruction and the content that the patient did not understand or remember well during the interaction with the system. 

In summary, our contributions are as follows: 
\begin{itemize}
    \item{We introduce a novel multi-component framework to help patient comprehend their Electronic Health Record (EHR) notes, more specifically discharge instruction. EHRTutor significantly outperforms GPT-4 with a simple prompt setting.}
    \item{Our human evaluation shows that our system can produce high-quality patient education conversations. Finally, we generated 5k synthetic data that can be used for in-house system training.
    }
    
\end{itemize}

\section{Related works}
\subsection{Patient education}
Enhancing patients' comprehension of their discharge instructions is of paramount importance \citep{FEDERMAN201814, kwon2022medjex, weerahandi2018predictors}. A deficiency in understanding can result in inadequate self-care at home. Previous research and projects have endeavored to educate patients through Question-Answer (QA) systems and create hospital course summaries \citep{FEDERMAN201814}. These prior efforts included the development of a question generation module and an answer verification module for the QA system. Subsequently, some works have shown that learning through dialogues and conversations can notably improve patients' understanding of their health conditions\citep{cai-etal-2022-generating, golinkoff2019language, sultan-etal-2020-importance, xu-etal-2022-fantastic, yao-etal-2022-ais, zhang-etal-2020-dialogpt}. This work goes a step further. It combines question generation and answer verification modules together and applies Reasoning-Acting (ReAct) to make the conversation educate patients better.

\subsection{Clinical question generation and answering}
Successful patient education relies on effective questions \citep{pylman202012}. The question generation problem has been previously explored through template-based \citep{10.1162/COLI_a_00206, fabbri-etal-2020-template, kaplan2010human} and seq2seq model \citep{du-cardie-2017-identifying, duan-etal-2017-question, kim2018improving, shwartz-etal-2020-unsupervised, sultan-etal-2020-importance}. Although models fine-tuned using generic text have shown promise in Human interaction \citep{brown2020language, ouyang2022training, shwartz-etal-2020-unsupervised}. Due to the concerns about the hallucination issue of neural language model \citep{maynez-etal-2020-faithfulness, pagnoni-etal-2021-understanding} caused by insufficient medical knowledge \citep{sung2021can, yao2022extracting, yao2022context}, question generation in the medical field remains challenging. In this paper, we proposed a template-based prompting with a verification step based on LLMs to restrict questions to our clinical scope.

\subsection{LLMs as cooperative agents}
Our research extends the recent developments that utilize two LLMs as collaborative agents to facilitate multi-round conversation generation between doctor and patient \citep{Panait2005CooperativeML}. During the experiment, we encountered a similar issue as previous work \citep{li2023camel}, in which the patient agent repeats the wrong answer. To address this issue, we applied the ReAct framework to our doctor agent in order to give a more informative response to guide patient agent. Additionally, EHRTutor can also be used for data augmentation. Existing works show that it is effective to utilize LLMs generating training data for tasks such as code summarization\citep{li2023feasibility}. Also, LLMs have shown their capability for data generation and annotation for some tasks in the absence of expensive human annotation data \citep{dai2022promptagator, ding2023gpt3, Gilardi_2023, zhou2023large}. In this paper, we can use our framework to generate synthetic patient education conversations for in-house system training.

\section{Methods}
\begin{figure*}[t] 
    \centering
    \scalebox{0.8}{
    \includegraphics[width=1.0\textwidth, height=5cm]{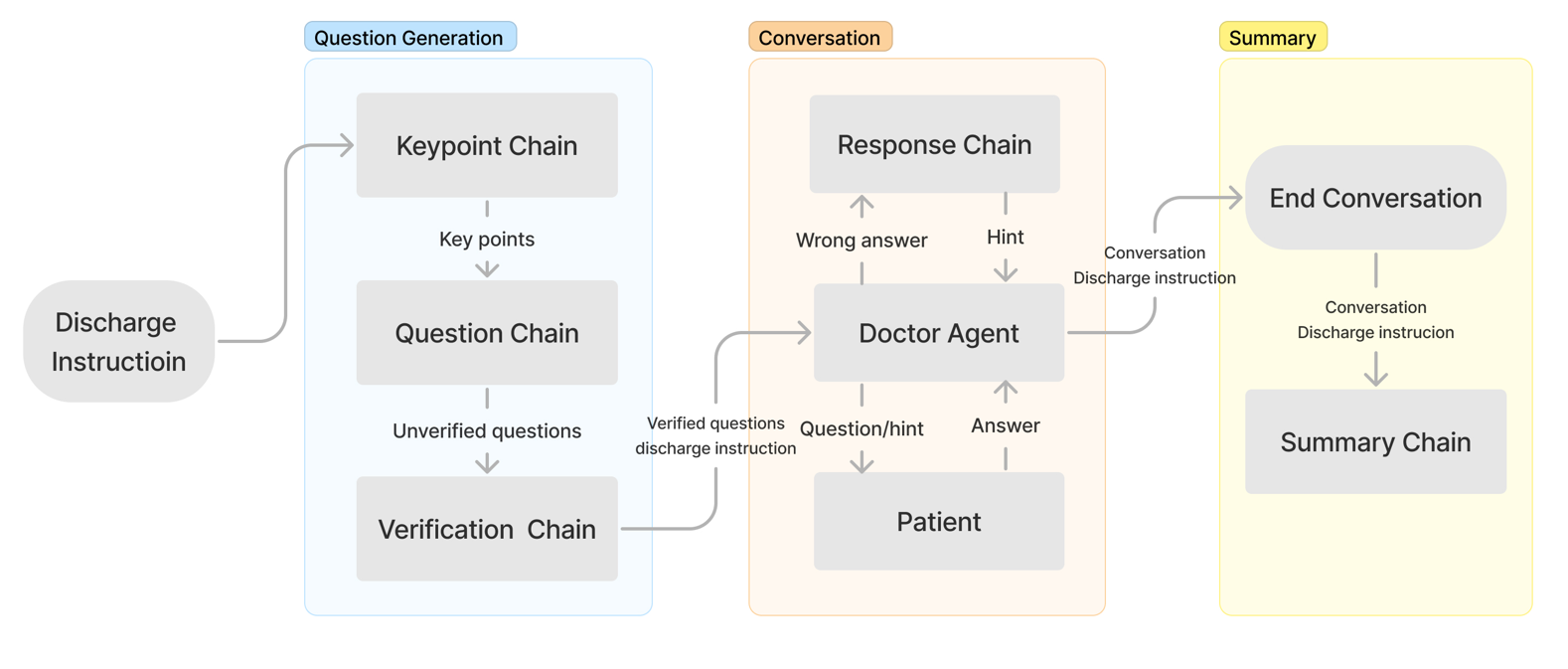}
    }       
    \caption{\textbf{The pipeline for EHRTutor} Here is an overview of the pipeline for EHRTutor. A more detailed explanation is provided in the method section.}
    \label{fig:agent_flowchart}
\end{figure*}

\textbf{Pipeline Description:}
Given a discharge instruction, the agent will generate questions based on the instruction. The doctor agent will ask those questions to the patient. According to the patient's response, the agent will decide which option to take: give a hint or ask the next question. Once the conversation is done, the agent will provide a summary to both the patient and the doctor as a conversation feedback. \\
\textbf{Keypoint chain}. The keypoint chain is used to generate the key points which will guide the question chain and hint chain to generate appropriate output. \\
\textbf{Question chain} The question chain is used to generate informative questions from discharge instruction using Large Language Model (LLM). Since zero-shot prompting will lead to generating inappropriate questions or meaningless questions. For example, what is the date you admit to hospital. Thus, we summarize the commonly asked clinical questions and split the questions into four general categories: $test$, $medication$, $complications \& progress$ and $follow up$ and made a question template (Table 1) for instruction prompting.  \\
\textbf{Verification chain}The verification chain follows the question chain. Since there are lots of questions that are meaningful but we cannot verify the answers of those questions based on discharge instruction. For example, questions related to the career or habit of the patient, the family disease history. Therefore, we add verification chain after the question chain to ensure we can verify the answer of every question we ask based on discharge instructions. \\
\textbf{Response chain} The response chain will be called by the agent as a tool when the agent thinks the patient’s answer is incorrect and needs a hint. The hint will be generated based on the discharge instruction and previous conversation. \\
\textbf{Agent}. The agent is role playing as a doctor who will ask patient questions and give hint based on the patient's response. Since the decision making need to based on the previous dialogue, we use the ReAct framework to implement the prompt of agent. We introduce the agent scrachpad in our prompt. For each decision, the reason of taking this action and the observation of the action will be saved in the agent scrachpad, which will be updated for every prompt. Previous decisions will also affect the next step. This dynamic process aids the agent in offering suitable hints or follow-up questions, enhancing patient education.\\
\textbf{Summary chain}. The summary chain takes the discharge instruction and conversation as the input and generates the summary based on our designed template. The output summary contains the key points for self-care and the points the patient did not understand or remember during the conversation. 

\section{Experiment}
\subsection{Experiment data}
Following previous work \citep{cai2023paniniqa}, we established the discharge instruction from the MIMIC-III database \citep{johnson2016mimic}\footnote{Experiment data and evaluations are conducted using Microsoft Azure's GPT-4.}. 
MIMIC-III is a publicly available repository of de-identified health records of over 40,000 patients collected from the Beth Israel Deaconess Medical Center in Massachusetts.
\subsubsection{Human Evaluation Settings}
The goal of human evaluation is to have human domain experts evaluate whether these machine-generated questions and responses are comparable to human-crafted questions and responses. 
To do so, we recruited two medical practitioners\footnote{Two medical students with hospital internship experience.} and their tasks are 1) to read the discharge instructions and conversation and provide qualitative feedback on whether these machine-generated questions and educating responses are educationally effective to the patients, if not, how should they be improved. 2) provide ratings on conversations and summaries generated by different methods.

We used the GPT-4 as our baseline. Specifically, we developed a simple prompt setting:
\textit{You should role-play as a doctor. Your job is to test your patient's understanding of discharge instructions by asking questions. Give a hint if they got the wrong answer for the first time. Here is the discharge instruction: \{ instruction\}}. We performed this prompt on GPT4 models and compared the performance with EHRTutor. To generate conversation data, we developed an LLM-based agent to role-play as a patient to give a response to the models. The patient agent will randomly generate the correct answer, wrong answer, or irrelevant phrase. The experts will evaluate the conversations and summarise them based on the following designed metrics. The total points are 5 for each metric. For each inappropriate question, response, or decision in the conversation, deduct one point. 

\begin{figure*}[t] 
    \centering
    \scalebox{0.8}{
    \includegraphics[width=0.8\textwidth, height=3.8cm]{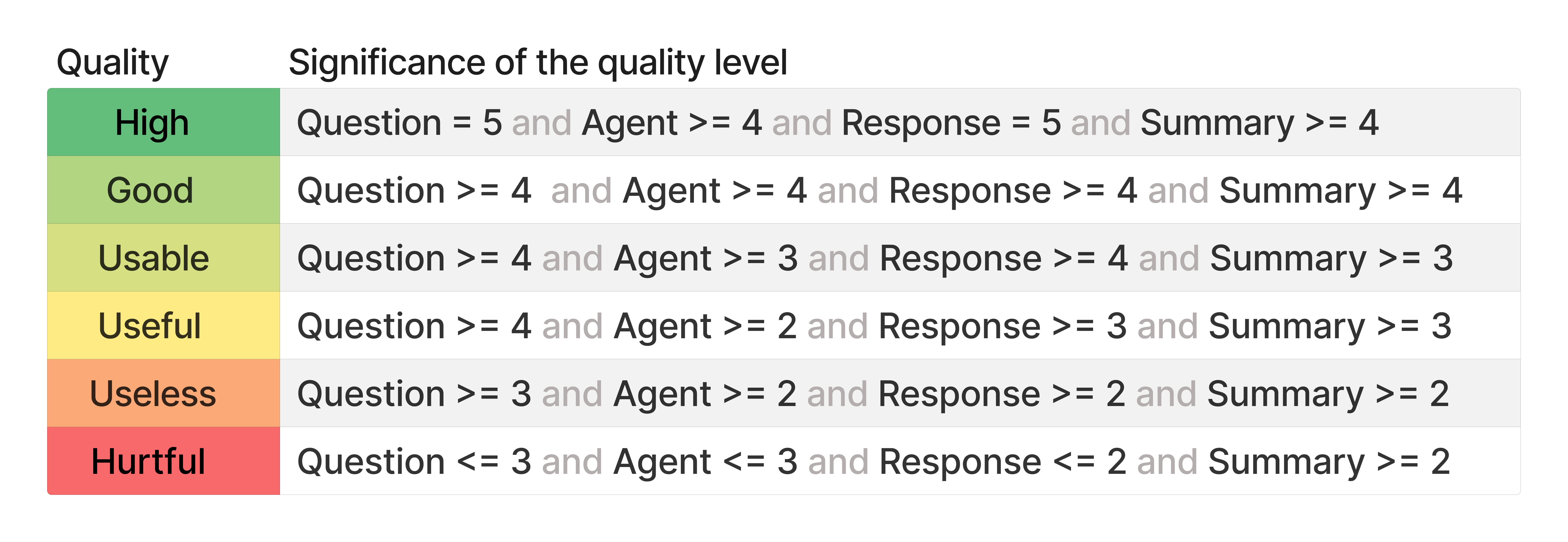}
    }
    \caption{The rules for classifying the model's overall performance}
    \label{fig:metrics}
    \vspace{-3mm}

\end{figure*}

\begin{figure*}[t] 
    \centering
    \includegraphics[width=1\textwidth, height=3.cm]{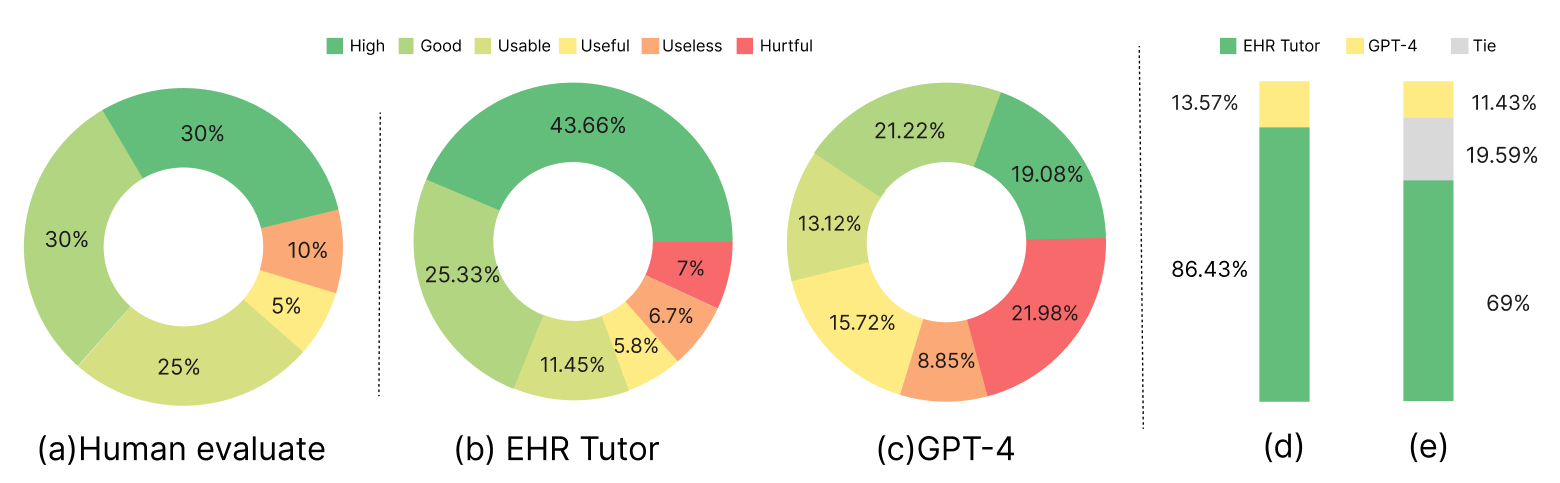}
    \caption{ \textbf{(a)} 
    Human evaluation quality level results for both EHRTutor and GPT-4 synthetic data. It demonstrates the practicality of using AI to generate synthetic data in this task.
    \textbf{(b)} Overall performance of EHRTutor generated conversation and summary based on LLMs evaluation. \textbf{(c)} Evaluation of GPT-4 generated conversation and summary based on LLMs evaluation.  \textbf{(d)} LLM preference between EHRTutor and GPT4 synthetic data\textbf{(e)} We rate the quality of each model's output based on Figure~\ref{fig:metrics} and compute the win rate.}
    \label{fig:pie_charts}
    \vspace{-3mm}
\end{figure*}
\textbf{Question}. We will examine the questions in the following metrics: 1) Cover rate. Do the questions cover the test, medication, complications \& progress in the discharge instruction? 2) Relevance. Is the question relevant to the discharge instruction? 3) Fluent. A concise and clear syntax and vocabulary, devoid of unnecessary questions.
\textbf{Response}. We will examine the response in the following metrics: 1) Relevance. The response should be pertinent to the question and the patient’s initial answer; however, it should not directly provide the correct answer upon the patient’s first incorrect response. 2) Sufficient. The response can help the patient recall or comprehend the correct answer. 3) Factuality. Responses should align with the medical facts and content in the discharge instructions. 
\textbf{Agent}. We will examine the agent based on the decision-making process. The bad decision includes incorrect assessment of the patient's answer, asking repeated questions, and ending the conversation early.
\textbf{Summary}. The summary should answer the following questions(if applicable): 1) What is the patient's medical problem? 2) What are the patient's medical allergies? 3) What exercises are good for the patient? 3) What should the patient eat? 4) What activities should the patient avoid? 5) When is the patient's next appointment? 6)What are the points that patients did not understand or memorize well?
\subsubsection{LLMs evaluation settings}

Following the recent practice of applying large language models in evaluating dialogue tasks \citep{liu2023gpteval}, we utilize two well-known SOTA LLMs (GPT-4, Claude2) as the evaluation model to measure the quality of patient education automatically. Since the performance of ChatGPT is worse than GPT-4, we did not choose ChatGPT for evaluation.
Similar to the human evaluation in the previous section, we evaluate the quality of generated conversations and summaries based on the same metrics as human evaluation in four perspectives (i.e., \emph{Question}, \emph{Response}, \emph{Agent}  and \emph{Summary}). We collect the evaluation model's responses and report the score of each metric. To evaluate overall quality, we compute the average score for each perspective and rate the overall quality based on Figure ~\ref{fig:metrics}. Our prompt to the evaluation model is shown in Appendix~\ref{appendix:LLM_eval_prompt}.

\subsection{Experiment results}
\subsubsection{Human evaluation results}

The results of human evaluation is in Figure~\ref{fig:pie_charts}(a) \& Table~\ref{table:human_eval}. We also interviewed two medical practitioners to identify the high-quality and low-quality questions and responses from ten sets of conversation and summary data. The feedback from the physicians shows that both model, EHR Tutor and GPT4, performs well on the following aspects: 1) All questions and most responses are helpful for the patient to understand the discharge instruction. 2) All questions do not need medical background to answer. 3) Most responses are factually correct. However, there are still some aspects that need improvement. Both EHRTutor and GPT4 model exists in some cases that do not cover all key points in the discharge instruction. Some questions generated by GPT4 cannot be verified based on the discharge instructions. Take example 1 in Appendix~\ref{appendix:case_study} as an example. Although the discharge instruction mentioned that the patient is allergic to Aspirin, we cannot guarantee that the patient is not allergic to other medicines or food. Some questions may require the patient to remember too many medical terms (Appendix~\ref{appendix:case_study}  example 2). 
This may contain more than ten symptoms. We can change the question by listing examples and asking the patient to answer the rest. 

\begin{table}
    \centering
    \scalebox{0.85}{
    \begin{tabular}{|c|c|c|c|c|c|c|c|c|}
        \cline{1-8}
          \multicolumn{3}{|c|}{Question} & \multicolumn{1}{c|}{Agent} & \multicolumn{3}{c|}{Response} & \multicolumn{1}{c|}{Summary}\\
        \cline{1-8}
          Cover rate & Relevance & Fluent & Rationality & Relevance & Sufficient & Factuality & Cover rate\\
        \hline
         4.80 & 4.85 & 4.90 & 4.36 & 5 & 4.8 & 5 & 4.43\\
        \hline
    \end{tabular}
    }
    \caption{Human evaluation results for each feature.}
    \label{table:human_eval}
    \vspace{-4mm}
\end{table}

\subsubsection{LLMs Evaluation results}

\begin{table}
    \centering
    \scalebox{0.8}{
    \begin{tabular}{c|c|c|c|c|c|c|c|c|}
        \cline{2-9}
         & \multicolumn{3}{c|}{Question} & \multicolumn{1}{c|}{Agent} & \multicolumn{3}{c|}{Response} & \multicolumn{1}{c|}{Summary}\\
        \cline{2-9}
         & Cover rate & Relevance & Fluent & Rationality & Relevance & Sufficient & Factuality & Cover rate\\
        \hline
        \multicolumn{1}{|c|}{EHRTutor} & 4.59 & 4.80 & 4.35 & 4.11 & 4.64 & 4.35 & 4.75 & 4.32\\
        \hline
        \multicolumn{1}{|c|}{GPT-4} & 3.36 & 3.60 & 3.39 & 2.86 & 3.11 & 2.81 & 3.42 & 3.78\\
        \hline
    \end{tabular}
    }
    \caption{LLM evaluation results for each feature.}
    \label{table:evaluation_LLM}
    \vspace{-6mm}
\end{table}
We utilize Language Model Models (LLMs), specifically GPT-4 and Claude-2, to assess our conversation and summary quality using the same metric as human evaluation. Quality ratings are determined based on Figure~\ref{fig:metrics}. We conduct a comparative analysis of EHRTutor and GPT-4, calculating the win rate as depicted in Figure~\ref{fig:pie_charts}e. Additionally, we solicit preferences from the LLMs for output between EHRTutor and GPT-4, as indicated in Figure~\ref{fig:pie_charts}d. These results consistently align with the human evaluation results, as depicted in Figure~\ref{fig:pie_charts}b and Figure~\ref{fig:pie_charts}c. Furthermore, the average scores across various metrics, as presented in Table~\ref{table:evaluation_LLM}, demonstrate EHRTutor's outperforms compared to GPT-4.

\section{Limitation}
Due to the potential for hallucinations in Language Model Models (LLMs), the evaluation results of LLMs may exhibit bias. Furthermore,  Although EHRTutor shows a strong ability in question generation and response to the patient, there still is a possibility of providing incorrect or harmful information to patients. For example, LLM may assume the patient only got the symptoms mentioned in the discharge instruction and consider the answer related to symptoms not shown in the instruction as incorrect. 

\section{Conclusion}
In this study, we introduce EHRTutor, an LLM-based patient education framework. The EHRTutor generates educational questions based on discharge instruction and educates patients through conversational question-answering. Human and LLMs evaluation shows that EHRTutor has better performance than the baseline.

\newpage


\bibliographystyle{plainnat}
\bibliography{neurips_2023}

\newpage

\appendix

\section{Question generation template}
\begin{table}[h]
    \centering
    \begin{tabularx}{\textwidth}{X}

    \toprule
    \textbf{Test}\\
    1. What is the [test] for?\\
    2. Do you understand the purpose of the [test/procedure] performed during your hospital stay?\\
    3. How many times have you done this [procedure]?\\
    \midrule
    
    \textbf{Medications}\\
    1. What is the frequency of taking this [medicine]? \\
    How and when will you take your prescribed medications?\\
    2. How long will you be taking these medications? \\
    Which medications have had a change in how you take them? \\
    3. Why do you need this [treatment]?\\
    4. Are there any side effects for [medicine]?\\
    5. Have you been informed about potential side effects or complications associated with your prescribed medications or treatments?\\
    6. Will this [medicine] interact with medicines that I'm already taking?\\
    \midrule
    
    \textbf{Complications and progress} \\
    1. What are the possible complications?\\
    2. What signs should you be alert to regarding your condition?\\
    3. What steps should you take if you experience these symptoms?\\
    4. Do you know how to recognize and monitor any warning signs or symptoms related to your condition at home?\\
    \midrule
    
    \textbf{Follow up}\\
    1. Have you been provided with information about when and how to schedule follow-up appointments with your healthcare provider?\\
    \bottomrule
    
    \end{tabularx}
    \vspace{0.05in}
    \caption{Question generation template. We categorize the problems into four groups and offer various ways and perspectives for asking questions in each category.}
\end{table}

\begin{table}[ht]
    \centering
    \begin{tabularx}{\textwidth}{X}

    \toprule
    \textbf{1. Provide a response to a question that cannot be verified.}\\
    \textbf{Discharge instruction}: "...You also had an aspirin allergy. You were evaluated..." \\
    \textbf{Doctor}: Can you tell me what you're allergic to?\\
    \textbf{Patient}: I'm allergic to peanuts.\\
    \textbf{Doctor}: That is incorrect. You're not allergic to peanuts but to aspirin. However, you have undergone aspirin desensitization during your treatment. \\
    \midrule

    \textbf{2. The answer requires too many element for patients to remember.}\\
    \textbf{Discharge instruction}: "...Please call your doctor or go to the ER if you develop any new neurologic symptoms, such as headache, weakness, numbness, slurred speech, confusion, dizziness, or trouble walking, as these could be signs of another stroke." \\
    \textbf{Doctor}: What are the potential signs of another stroke that you should watch out for?\\
    \textbf{Patient}: I only remember headache, weakness.\\
    \bottomrule
    
    \end{tabularx}
    \vspace{0.05in}
    \caption{The case studies}
    \label{table:case_study}
\end{table}
\label{appendix:case_study}

\section{Prompt for LLM evaluation}

    The previous tasks is educate patient through conversation. We generated three conversation using different models,
    EHRTutor and GPT4. Your job is to compare the performance of each model based on the following metrics.
    The full point is 5 for each metric. For each sentence did not satisfy the metric deduct one point. 
  
    Question:
    1. Coverrate: Does the question covers all keypoints of discharge instruction?
    2. Relevance: Do the questions relevance to discharge instruction?
    3. Fluent: A concise and clear syntax and vocabulary, devoid of unnecessary question 
    
    Agent:
    1. Rationality: Does the agent make correct decision? the wrong decision including incorrect assessment of patient's
      answer, end the conversation without ask all questions.
  
    Response:
    1. Relevance: Is the response relevant to the question and patient's answer
    2. Sufficient: Can the patient recall or comprehend the correct answers based on hints?
    3. Factuality: Is the response align with medical fact
    
    Discharge instruction:
    \{instruction\}
  
    EHRTutor:
    \{EHRTutor\}
  
    GPT4:
    \{GPT4\}
  
    Your output should be a dictionary object, adhering to the following structure:
    \{\{"best model": "the model with best performance, should be one of EHRTutor, GPT4", "EHRTutor": \{\{"Question": \{\{"Coverrate": "score", "Relevance": "score", "Fluent": "score"\}\}, "Agent": \{\{"Correctness": "score"\}\}, "Response": \{\{"Relevance": "score", "Sufficient": "score", "Factuality": "score"\}\}\}, "GPT4": \{\{"Question": \{\{"Coverrate": "score", "Relevance": "score", "Fluent": "score"\}\}, "Agent": \{\{"Correctness": "score"\}\}, "Response": \{\{"Relevance": "score", "Sufficient": "score", "Factuality": "score"\}\} \}\}\}\} 
    \label{appendix:LLM_eval_prompt}

\end{document}